\pdfoutput=1

\documentclass[11pt]{article}

\usepackage[preprint]{acl}

\usepackage{times}
\usepackage{latexsym}

\usepackage[T1]{fontenc}

\usepackage[utf8]{inputenc}

\usepackage{microtype}

\usepackage{inconsolata}

\usepackage{graphicx}

\usepackage{amssymb}
\usepackage{makecell}
\usepackage{float}

\definecolor{mypurple}{RGB}{103, 78, 167}

\title{Accelerating Antibiotic Discovery with Large Language Models and Knowledge Graphs}

\author{
\textbf{Maxime Delmas\textsuperscript{1}},
\textbf{Magdalena Wysocka\textsuperscript{1,2}},
\textbf{Danilo Gusicuma\textsuperscript{1}},
\textbf{André Freitas\textsuperscript{1,2,3}}
\\
\\
\textsuperscript{1}Idiap Research Institute, Switzerland \\
\textsuperscript{2}National Biomarker Centre (NBC), CRUK Manchester Institute, UK \\
\textsuperscript{3}Department of Computer Science, Univ. of Manchester, United Kingdom \\}

\begin{document}
\maketitle
\begin{abstract}
The discovery of novel antibiotics is critical to address the growing antimicrobial resistance (AMR). However, pharmaceutical industries face high costs (over \$1 billion), long timelines, and a high failure rate, worsened by the rediscovery of known compounds. We propose an LLM-based pipeline that acts as an alarm system, detecting prior evidence of antibiotic activity to prevent costly rediscoveries. The system integrates organism and chemical literature into a Knowledge Graph (KG), ensuring taxonomic resolution, synonym handling, and multi-level evidence classification. We tested the pipeline on a private list of 73 potential antibiotic-producing organisms, disclosing 12 negative hits for evaluation. The results highlight the effectiveness of the pipeline for evidence reviewing, reducing false negatives, and accelerating decision-making. The KG for negative hits and the user interface for interactive exploration will be made publicly available at \url{Anonymous GitHub}.

\end{abstract}

\section{Introduction}

Antibiotics are naturally occurring chemical compounds produced by organisms, known as natural products, that can inhibit the growth or eliminate bacteria and other microorganisms \cite{waksman_what_1947}. However, the introduction, use, and overuse of new antibiotics inevitably lead to the emergence of resistant pathogens \cite{altarac_challenges_2021}, and Antimicrobial Resistance (AMR) has been recognized as one of the top ten global public health threats \cite{eclinicalmedicine_antimicrobial_2021}. This ongoing cycle drives a continuous race to expand the antibiotic spectrum and treat patients infected with multidrug-resistant pathogens (MRPs) \cite{ahmed_antimicrobial_2024, iskandar_antibiotic_2022}.

\begin{figure}[ht]
\centering
  \includegraphics[width=0.8\columnwidth]{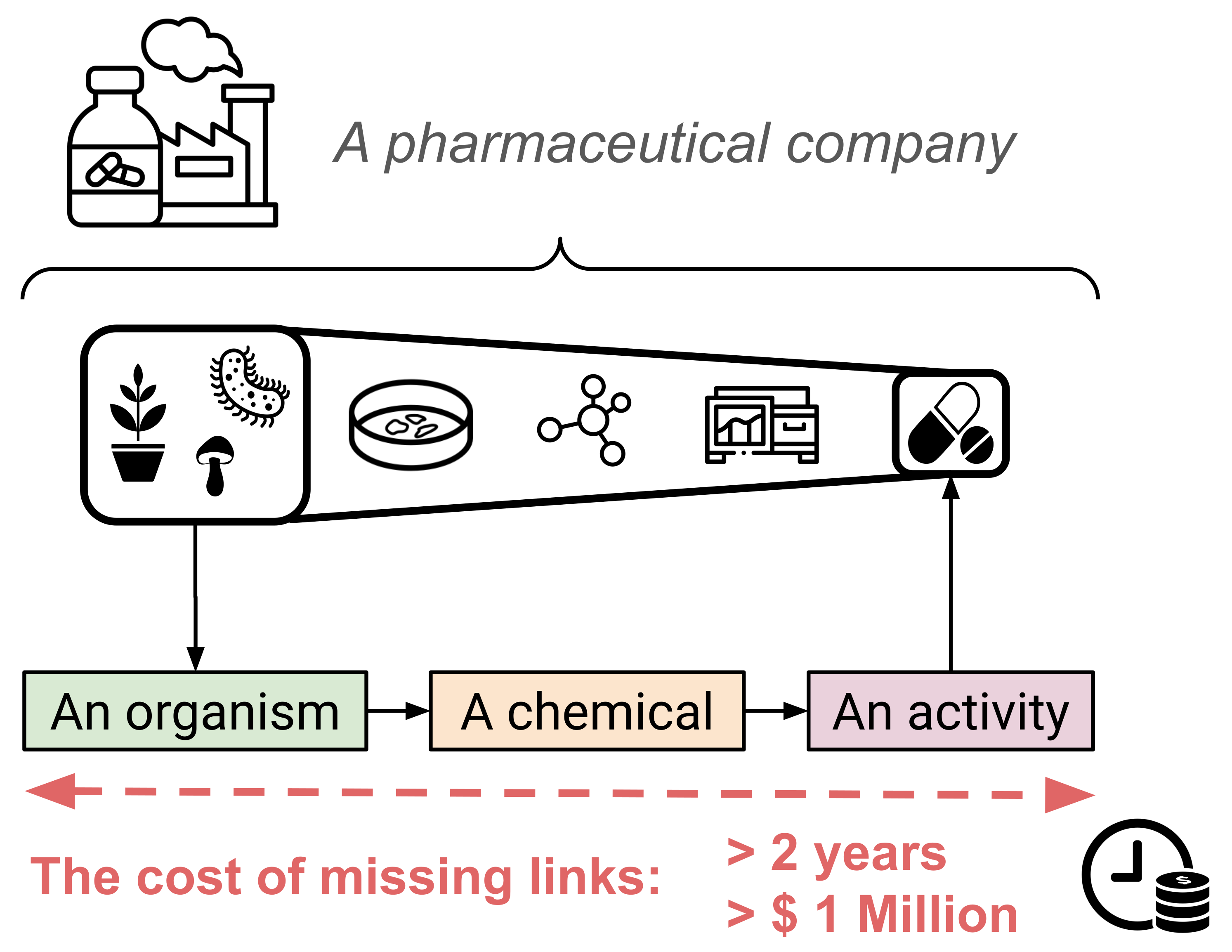}
  \caption{An overview of the early phase of antibiotic development and the cost attached to lead compounds identification.}
  \label{fig:fig1}
\end{figure}

The development of new antibiotics is highly challenging \cite{payne_drugs_2007, altarac_challenges_2021}. The process has a high failure rate, and the total cost from identifying lead compounds to market approval can exceed \$1 billion and take over a decade \cite{ardal_antibiotic_2020, wouters_estimated_2020}. In the initial phase, pharmaceutical companies explore ecosystems \cite{quinn_going_2024}, searching for exotic organisms that produce novel bioactive compounds (see Figure \ref{fig:fig1}). This phase involves identifying and isolating these compounds and evaluating their activity against MRPs. Identifying promising lead compounds (those with the highest potential for success) can already require over \$1 million and years of research \cite{ardal_insights_2018}. A major challenge in this early phase is avoiding rediscovery scenarios, when a potentially active compound has already been reported in scientific literature or patent databases. Such prior knowledge often eliminates the compound's commercial value by removing its novelty. In addition, one can consider that if an active molecule produced by an organism is publicly known but not already commercialized, it is likely that it has already been tested but failed in later clinical stages. Therefore, ensuring comprehensive awareness of existing research is critical to avoid costly investments in non-viable targets. As stated by \citep{paul_how_2010}, if a candidate has to fail, it is critical to it make fail faster and less expensively.

Preventing rediscoveries requires an extensive review of scientific literature, databases, and patents related to the initial list of target organisms. This task is firstly complicated by the unstable taxonomy and nomenclature of organisms \cite{beninger_understanding_2019}. Many organisms have been repeatedly rediscovered and reclassified under different names. For instance, \textit{Cephalosporium acremonium}, \textit{Hyalopus acremonium}, \textit{Acremonium strictum} and \textit{Sarocladium strictum}, published in 1882, 1941, 1971 and 2011 respectively, all refer to the same organism under the most recent classification. To capture relevant data, literature reviews must expand the search for such synonyms.

Evidence of prior activity can appear in diverse forms. Some references from the literature of the organism describe its activity without identifying specific active compounds, e.g., “The culture of A inhibited the growth of \textit{Staphylococcus aureus}." Others may report the isolation of a compound from the organism without detailing its biological activity ("Compound C was isolated from organism A"), requiring a 2-hop search for chemical activity evidence (e.g. Compound C exhibited antibacterial activity against \textit{Staphylococcus aureus}.")

This review process is traditionally manual and extremely time-consuming. \cite{allen_estimating_1999} previously estimated over 1,000 hours to review 2,500 citations. There is a need for semi-automation given the expanding scientific literature and the high cost of false negatives. In this context, large language models (LLMs) have emerged as powerful tools for assisting literature reviews, particularly in the biomedical domain \cite{wysocka_large_2024, yun-etal-2023-appraising, liao_llms_2024, hsu_chime_2024}. Beyond review, an effective solution would serve as an alarm system, flagging previously reported antibiotic activities associated with target organisms. Compared to novelty detection \cite{ghosal_novelty_2022}, we rather seek for non-novelty detection for relations between organisms, chemicals, and activities.

\begin{figure*}[ht]
  \centering
  \includegraphics[width=0.8\textwidth]{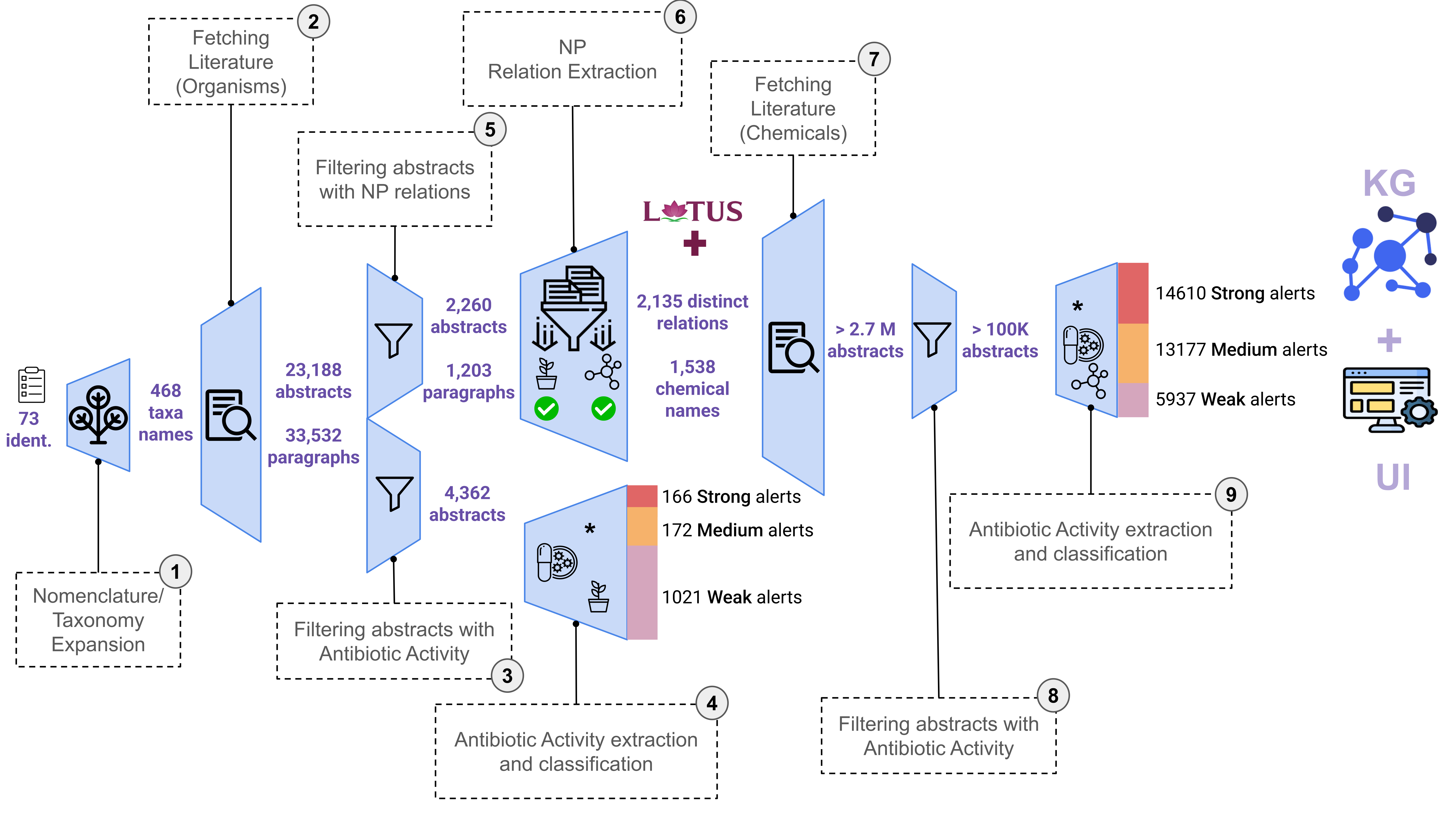}
  \caption{An illustration of the proposed pipeline, step-by-step, from the initial list of organism identifications to the extraction of AA evidence alerts in 3 levels. Intermediary annotations (in \textcolor{mypurple}{\textbf{purple}}) describe the flow of literature, relations, and evidence that have been processed.}
  \label{fig:fig-pipeline}
\end{figure*}

In this work, we propose an LLM-based pipeline to automate the construction of such an alarm system. The system is based on a Knowledge Graph (KG), ensuring taxonomic and nomenclature resolution, interoperability between natural product resources, and classification of evidence into three alert levels. We demonstrate this system in a real industrial setting with a private input list of 73 organisms and disclosing 12 on which we evaluated the system's performance. We intend to release the associated KG and the user interface for exploring evidence and alerts.

\section{Data}

Our dataset is composed of an initial private list of 73 organism identifications, from which we disclosed 12 negative hits for evaluation after evidence of already reported activity have been found. This review was conducted by a team of three experts, using public literature (PubMed), databases (LOTUS \cite{rutz_lotus_2022}) and proprietary tools (eg. CAS SciFinder \cite{gabrielson_scifinder_2018}). See details in appendix \ref{sec:manual-review}. For the proposed alert system, we excluded proprietary resources and decided to primary focus on two large public resources: PubMed and LOTUS. LOTUS is an open, community-curated database containing over 750,000 structure-organism pairs which is hosted on the Wikidata KG. Taxnonomic and nomenclature information of organisms are extracted from the GBIF backbone taxonomy \cite{registry-migrationgbiforg_gbif_2023}, a comprehensive and synthetic classification that integrates taxonomic data from multiple sources.

\section{Methodology}
\label{sec:methods}

This section provides a step-by-step description of the pipeline represented in Figure \ref{fig:fig-pipeline}. The input is a list of user-defined organism identifications. Identifications can be specific, at the species level (e.g., \textit{Aspergillus calidoustus}), or unspecific (represented by the abbreviation \textit{sp.}), indicating an undetermined species within a genus\footnote{A genus is a taxonomic rank grouping species that share common characteristics.} (e.g., \textit{Aspergillus sp.}).

In step \textit{(1)}, each identification is aligned with an entity in the GBIF taxonomy. Species-level identifications are expanded to include all known synonyms, while genus-level identifications are expanded to encompass all species within the genus and their respective synonyms. In step \textit{(2)}, abstracts and relevant paragraphs from PubMed full-text articles are retrieved using the NCBI EUtils API\footnote{\url{https://dataguide.nlm.nih.gov/eutilities/utilities.html}}.

Step \textit{(3)} filters the organism literature to exclude articles irrelevant for antibiotic activity (AA) evidence extraction (e.g., ecology, environmental studies, genetics). A lightweight lexical classifier, trained on MeSH\footnote{MeSH are standardized biomedical indexing terms in PubMed.} annotations, ensures efficient filtering. In step \textit{(4)} we prompt the LLM (Mixtral-8x7b \cite{jiang2024mixtral}) for Zero-shot extraction of AA evidence from the selected abstracts \cite{kojima2022large}. These evidence, derived solely from the organism's literature, are designated as \texttt{OL}-evidence (Organism-Literature). Evidence are then categorized into three alert levels: \texttt{Strong} (direct experimental evidence of activity), \texttt{Medium} (indirect, imprecise, or minor evidence), and \texttt{Weak} (no substantial evidence) using the LLM. More details about the prompting strategy and concrete examples in appendix \ref{app:AA-classifications}

Steps \textit{(5)} to \textit{(7)} focus on identifying chemicals isolated from the organisms. Similar to \textit{(3)}, step \textit{(5)} filters literature to retain only texts likely to report chemical isolations. Since MeSH annotations are unavailable for this task, we used LLM-generated pseudo-labels to train a second lexical classifier \cite{wang2023large}. Details on the classifiers used for filtering are provided in Appendix \ref{app:filtering-classifiers}.

In step \textit{(6)}, a natural products Relation Extraction (RE) model \cite{10.1162/coli_a_00520} (fine-tuned from BioMistral-7B \cite{labrak2024biomistral}) processes selected passages to extract natural product relations (\texttt{NPR}). These relations are sourced from abstracts (\texttt{TiabNPR}) or paragraphs (\texttt{ChunkNPR}), then augmented with relations from the LOTUS database (\texttt{LotusNPR}).

Steps \textit{(7)} to \textit{(9)} mirror steps \textit{(2)} to \textit{(4)}, but use the extracted chemical names as input. This produces a prioritized list of chemical literature evidence (\texttt{CL}-evidence), categorized into the same three alert levels.

\begin{figure*}[ht]
  \centering
  \includegraphics[width=0.7\textwidth]{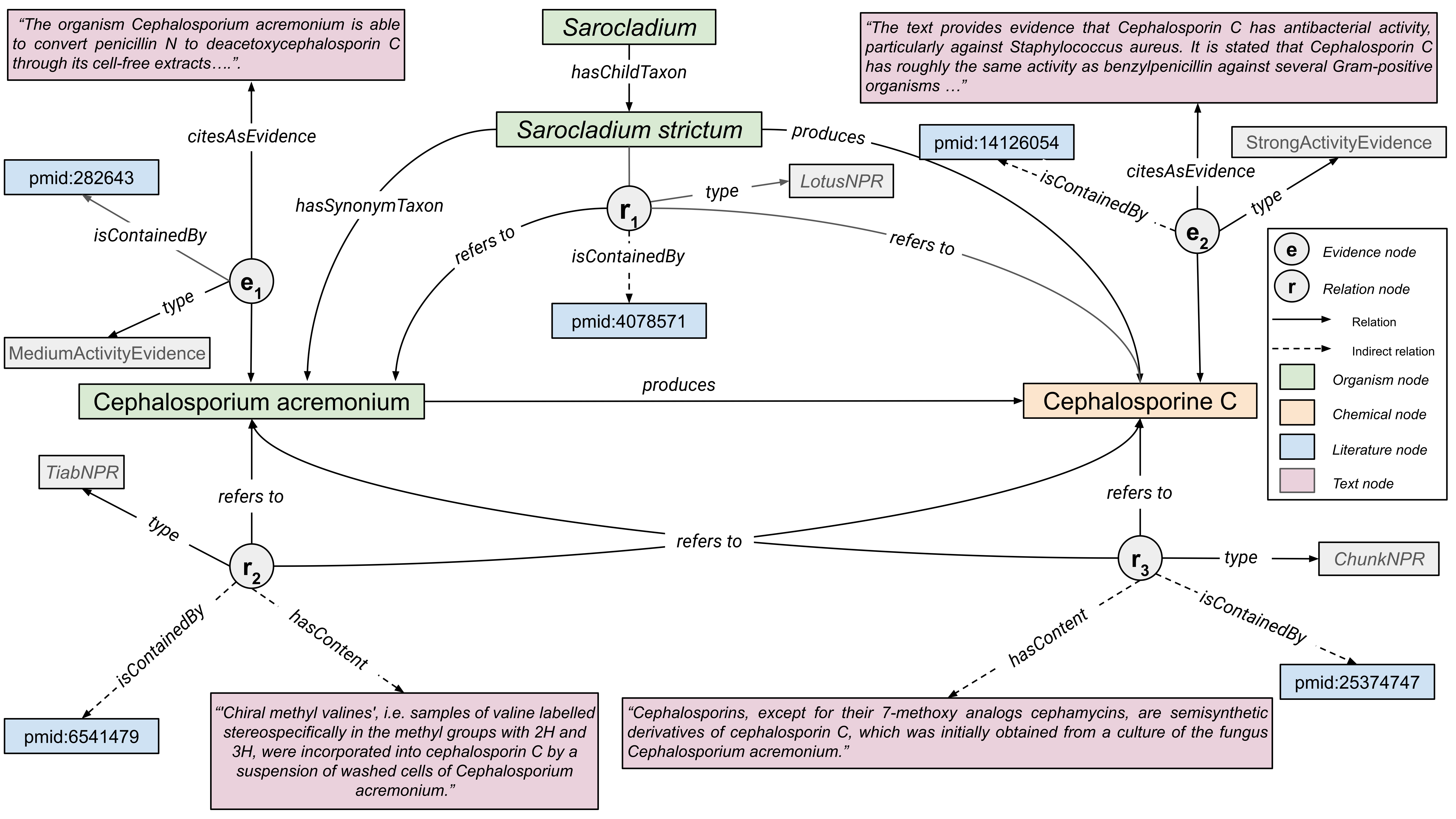}
  \caption{A snapshot of the built KG around the natural product relation between \textit{Cephalosporium acremonium} and Cephalosporine C. Taxonomic and nomenclature relationships are represented between Organism nodes in green. Relation nodes $(r_1, r_2, r_3)$ describe relations between organisms and the isolated natural product Cephalosprin C from different sources: LOTUS database (\texttt{LOTUSNPR}) and extracted from an abstract (\texttt{TiabNPR}) and a paragraph (\texttt{ChunkNPR}). Text nodes connected to relation nodes $(r_2, r_3)$ refer to the text from which the relation was extracted. The evidence node $e_1$ is an example of \texttt{OL}-evidence associated with a \texttt{Medium} alert. The node $e_2$ is a \texttt{CL}-evidence associated with a \texttt{Strong} alert. Literature node connected to relation and evidence nodes allow for linked to the original reference in PubMed (or using the DOI if not available in the case of LOTUS annotations).}
  \label{fig:fig-kg}
\end{figure*}

All processed data, including nomenclature, relations, literature, and alerts, are integrated into a Knowledge Graph (KG) using a dedicated ontology. Figure \ref{fig:fig-kg} provides a snapshot centred on the example of \textit{Sarocladium strictum} and its active metabolite \textit{Cephalosporin C}. The KG supports transparent resolution of taxonomic and synonym relations (e.g. \textit{Sarocladium strictum} \texttt{hasSynonymTaxon} \textit{Cephalosporium acremonium}), ensures interoperability between sources of relations (\texttt{LotusNPR}, \texttt{TiabNPR}, \texttt{ChunkNPR}), and, differentiates evidence origins (\texttt{OL} vs. \texttt{CL}) and alert levels (\texttt{Strong}, \texttt{Medium}, \texttt{Weak}).

\section{Results}

\subsection{Natural products literature:  descriptive bibliometric analysis}
\label{sec:bilbio}

\begin{figure}
  \centering
  \includegraphics[width=0.4\textwidth]{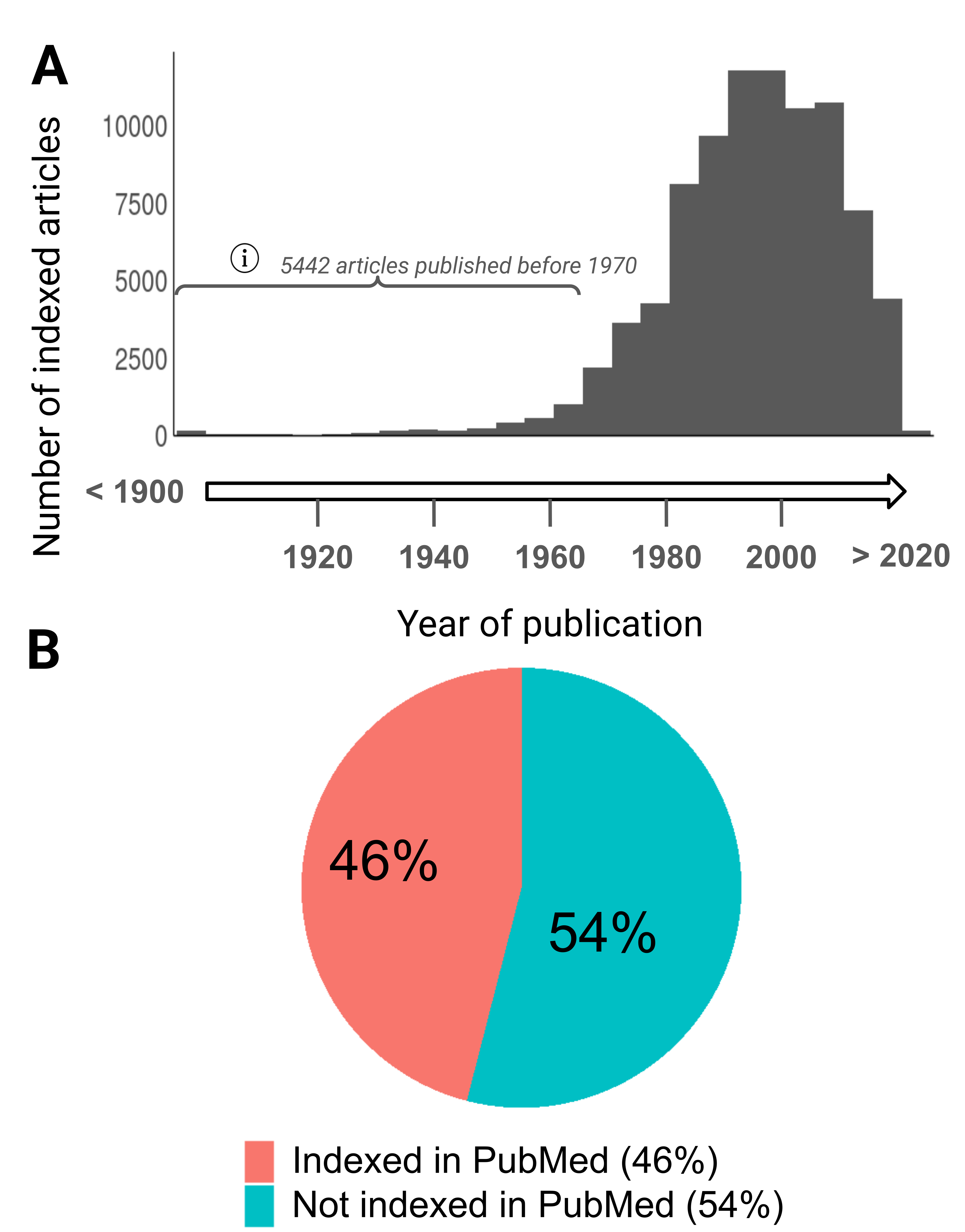}
  \caption{\textbf{A} describes the distribution of publication years for literature references annotated in the LOTUS database. Panels \textbf{B} represents the distribution of references indexed in PubMed for natural product relations annotated in LOTUS.}
  \label{fig:fig-lit-overview}
\end{figure}

Assessing the size and growth of the natural products and antibiotics literature is crucial to highlight the extensive effort required by reviewers. In 2024, it is more than 50,000 new articles that have been indexed in PubMed for the searches "natural products" and "antibiotics", reporting novel links between organisms, chemicals, and activities. While keeping up with new literature is crucial, Figure \ref{fig:fig-lit-overview}.A shows that a significant portion of annotated relations in the LOTUS database comes from older articles (pre-1970). Given the evolution of taxonomy and nomenclature over time, relying on original organism identifications from the text is unreliable, making synonym resolution essential for linking past and novel relations.
Using the publicly available literature from PubMed as a reference for an alert system also requires evaluating its coverage. Although PubMed includes over 38 millions articles, Figure \ref{fig:fig-lit-overview}.B indicates that fewer than half of the annotated references in the LOTUS database are actually indexed in PubMed. This observation underscores a notable gap in PubMed’s coverage. Nevertheless, given the extensive volume of literature within PubMed, it’s also reasonable to expect that many relevant references may be missing from LOTUS. Also, while we observed that most articles are in English, this likely reflect a bias from the resources used in LOTUS, and, other (eg. traditional Chinese medicine prescriptions) corpora are also expected to be relevant.

A notable example of the last points is Atranorin, an anti-inflammatory, analgesic, and antibacterial compound, isolated from \textit{Gyrophora esculenta} (now named \textit{Umbilicaria esculenta}), described in German by \citet{hoppe_galanthus_1958}.

\begin{figure*}
  \centering
  \includegraphics[width=1\textwidth]{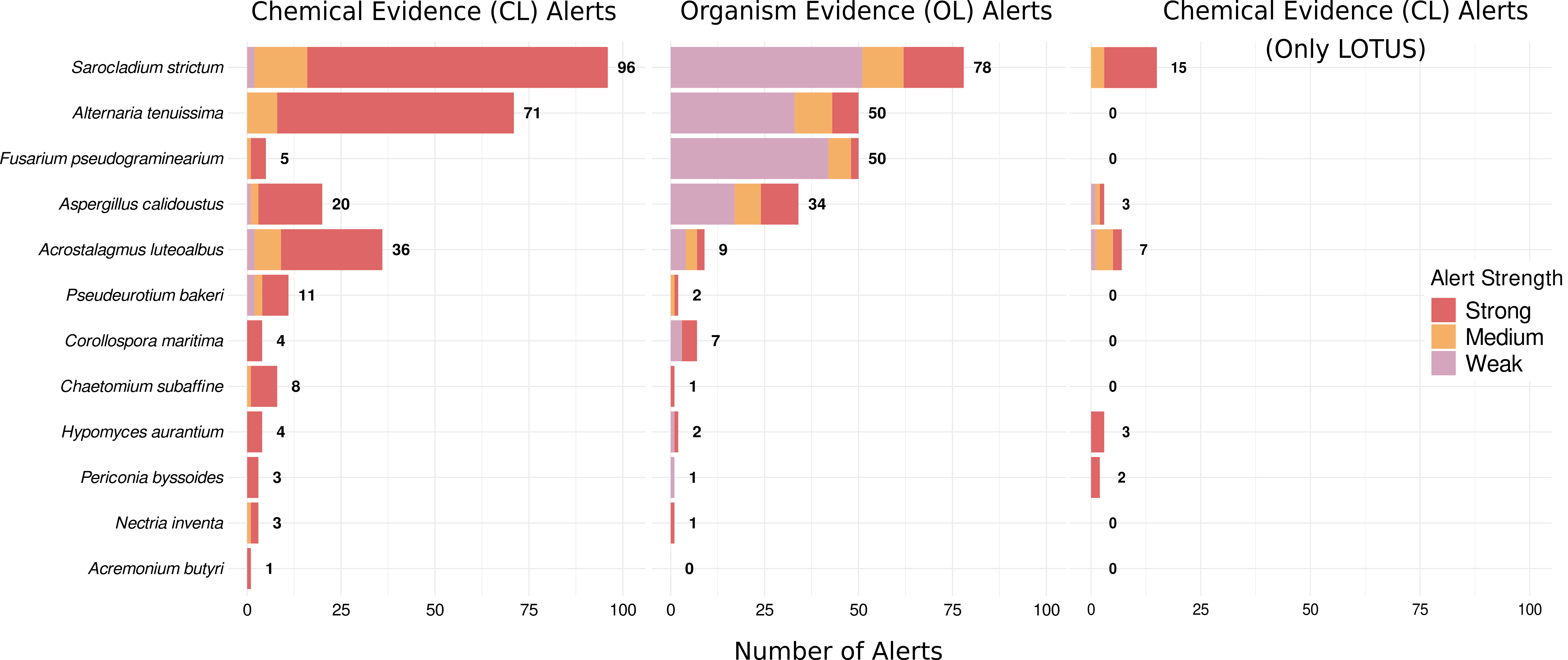}
  \caption{Distribution of all reported alerts per class (\texttt{Strong}, \texttt{Medium} and \texttt{Weak}) and categories for \texttt{CL} (left) and \texttt{OL} (center) evidence for the 12 discarded organisms. The right panel describes the reported evidence only using the LOTUS available natural products relations.}
  \label{fig:hits-alerts}
\end{figure*}

\subsection{Pipeline execution}

Starting from 73 initial identifications, the flow of extracted and processed literature is outlined in Figure \ref{fig:fig-pipeline}. Over 50,000 paragraphs were processed, yielding 2,135 organism-chemical relations and 1,359 alerts directly from the literature of organisms. Expanding to the literature of identified chemicals, more than 2.7 million abstracts were processed, resulting in 33,724 alerts for potential antibacterial activity.\footnote{Many alerts stem from genus-level identifications, which expand to numerous species.}

\begin{table*}[ht]
\centering
{\small
\begin{tabular}{llllll}
\hline
\textbf{Organisms} & \textbf{Chemicals} & \textbf{\makecell{PubMed ID\\Isolation}} & \textbf{\makecell{PubMed ID\\Activity}} & \textbf{RE / LOTUS} & \textbf{CL-evidence} \\
\hline
\textit{A. butyri} & Orbuticin & 8982351 & 8982351  & \checkmark/\checkmark & \textcolor{red}{Missed} \\
\textit{A. luteoalbus} & Acrozine A-C & 31226467 & 31226467 & \checkmark/\checkmark & \texttt{Strong} \\
\textit{A. luteoalbus} & T988 C & 35621985 & 35621985 & \texttimes/\texttimes & \textcolor{red}{Missed} \\
\textit{A. luteoalbus} & Lasiodipline E & 37627256 & 24529576 & \checkmark/\texttimes & \texttt{Strong} \\
\textit{A. luteoalbus} & luteoalbusin A & 23079524 & 35621985 & \checkmark/\checkmark & \textcolor{red}{Missed} \\
\textit{A. tenuissima} & Altertoxin I, II, III & 25260957 & 37764307 & \checkmark/\texttimes & \texttt{Strong} \\
\textit{A. tenuissima} & Tenuazonic acid & 34575812 & 34575812 & \checkmark/\texttimes & \texttt{Strong }\\
\textit{A. tenuissima} & Alternariol mono. ether & 24071643 & 38470179 & \checkmark/\texttimes & \texttt{Strong} \\
\textit{A. calidoustus} & Ophiobolin K & 25812930 & 29375031 & \checkmark/\texttimes & \texttt{Strong} \\
\textit{A. calidoustus} & Strobilactone A & 8698631 & \textit{ext. ref(1)} & \texttimes/\checkmark & \textcolor{red}{Missed} \\
\textit{S. strictum} & Cephalosporin C & 10397815 & 14126054 & \checkmark/\checkmark & \texttt{Strong} \\
\textit{S. strictum} & Isopenicillin N & 575040 & 7107525 & \checkmark/\checkmark & \texttt{Strong} \\
\textit{S. strictum} * & Cytosporone E & 29354097 & 22690142 & \checkmark/\texttimes & \texttt{Strong} \\
\textit{C. subaffine} & Chrysophanol & 35761187 & 25821480 & \checkmark/\texttimes & \texttt{Strong} \\
\textit{C. maritima} & Corollosporine & 16557326 & 16557326 & \checkmark/\texttimes & \texttt{Strong} \\
\textit{F. pseudograminearum} & Deoxynivalenol & 35878241 & 38408410 & \checkmark/\texttimes & \texttt{Strong} \\
\textit{F. pseudograminearum} & Zearalenone & 24291181 & 37929585 & \checkmark/\texttimes & \texttt{Strong} \\
\textit{H. aurantius *} & Cladobotryal & 9586194 & 12934912 & \texttimes/\checkmark & \texttt{Strong} \\
\textit{H. aurantius *} & Furopyridine antibiotics & 11918067 & 11918067 & \checkmark/\texttimes & \texttt{Strong} \\
\textit{H. aurantius} & Hypomycetin & \textit{ext. ref(2)} & \textit{ext. ref(2)} & \texttimes/\checkmark & \textcolor{red}{Missed} \\
\textit{N. inventa} & Chaetocin & 31569621 & 21140472 & \checkmark/\texttimes & \texttt{Strong} \\
\textit{N. inventa} & Verticillin B & 31569621 & 31569621 & \checkmark/\texttimes & \textcolor{red}{Missed} \\
\textit{P. byssoides} & Pericosine A & 18043803 & 26928999 & \checkmark/\checkmark & \texttt{Strong} \\
\textit{P. byssoides} & Macrosphelide A & 15895526 & 19298513 & \texttimes/\checkmark & \texttt{Strong} \\
\textit{P. bakeri} & Cytochalasin X & 35841670 & 35841670 & \checkmark/\texttimes & \texttt{Strong} \\
\textit{P. bakeri} & Chaetoglobosin B & 36104717 & 26669098 & \checkmark/\texttimes & \texttt{Strong} \\
\textit{P. bakeri} & Chaetoglobosin A & 36104717 & 26669098 & \checkmark/\texttimes & \texttt{Strong} \\
\hline
\end{tabular}
}
\caption{Comparison of reviewers extracted \texttt{CL}-evidence and system-extracted evidence for each discarded hits. When an organism is marked with a *, it indicates that the chemical has been retrieved for a synonym (eg. \textit{Cladobotyryum varium} in the case of \textit{Hypomyces aurantius}). "PubMed ID Isolation" and "PubMed ID Activity" list PubMed references for chemical isolation and antibiotic activity extracted by reviewers. The "RE/LOTUS" column uses a tick (\checkmark) and a cross (\texttimes) to show whether the relationship organism-chemical is present or missing. The left symbol represents extraction from the Relation Extraction (RE) pipeline, while the right symbol indicates whether it is annotated in the LOTUS database. \texttt{CL}-evidence indicates the system’s alarm level (\texttt{Strong}, \texttt{Medium}, \texttt{Weak}, or \textcolor{red}{Missed}). Ext. ref(1) and ext. ref(2) are non-PubMed references: \texttt{doi:10.1515/znb-2007-1218} and \texttt{10.3891/acta.chem.scand.51-0855}.}
\label{tab:hits-stats}
\end{table*}

\subsection{Evaluation on Discarded Hits}

Among the 73 initial identifications, 12 were discarded as negative hits after an extensive manual review. Figure \ref{fig:hits-alerts} displays the distribution of alerts raised for each discarded organism from \texttt{CL} (left) and \texttt{OL} (center) evidence. While the number of alerts varies (max: 174, min: 1), each organism has at least one \texttt{Strong} alert. To assess the impact of the extraction pipeline, an ablation study (Figure \ref{fig:hits-alerts} right) using only LOTUS database annotations showed that only 5 of the 12 negative hits could be identified, highlighting the added value of the RE step.
For the 12 negative hits, the reviewers previously identified 27 evidence triples (\textit{organism-chemical-activity}). Table \ref{tab:hits-stats} compares these with system-generated alerts from Figure \ref{fig:hits-alerts}, focusing on chemical-based alerts, as all evidence provided by the reviewers are linked to a chemical. An alert is considered missed if the chemical was not retrieved (via RE or LOTUS) or its activity was not reported\footnote{Neither \texttt{Strong}, \texttt{Medium}}. Among the 27 reviewer-reported evidence, 6 were missed by the system, including 3 because of non-indexed references or unavailable texts in PubMed. Notably, 26 of the 27 chemicals were successfully retrieved, with 22 through the RE step. A detailed error analysis is provided in Appendix \ref{sec:errorAnalysis}. 
Except for \textit{Acremonium butyri}, all negative hits were correctly discarded. Screenshots of the user interface, including an example for \textit{Sarocladium strictum}, are shown in Appendix \ref{sec:screenshots}.

\section{Discussion}

Most alert-associated chemicals were extracted from the public literature, suggesting an underestimation of PubMed's coverage in section \ref{sec:bilbio}, and, highlighting gaps in public databases, particularly for rarely mentioned organisms. However, given the nature of the task, and the cost of false negatives (e.g., \textit{Acremonium butyri}), public resources alone are insufficient to prevent rediscoveries. Notably, half of the missing evidence could have been recovered by incorporating non-publicly accessible literature, beyond PubMed and LOTUS.
From the initial set of 73 organisms, over 35,000 alerts were generated, which, paradoxically, could overwhelm the reviewers. To mitigate this, the prioritization system, categorizing evidence into \texttt{Strong}, \texttt{Medium}, and \texttt{Weak}, is essential for the reviewing process. Interestingly, in only 9 of the 27 evidence reported by the annotators, the activity of the chemical was reported in the same article as its isolation. This highlights the need for extending the search to the literature of individual chemicals, and reflects the 2-hop nature of the task. Moreover, accurate nomenclature resolution, inherently supported by the KG, remains critical. This is exemplified by the case of \textit{Hypomyces aurantius}, where key evidence were retrieved under its synonym \textit{Cladobotryum varium}. While a single (\texttt{Strong}) evidence is enough to discard an organism, comparing Table \ref{tab:hits-stats} and Figure \ref{fig:hits-alerts} suggests that many pieces of evidence may have been overlooked by reviewers, considering the vast amount of literature to examine. Paradoxically, in the proposed scenario, a "positive" result is therefore an "empty" result, such that no external evidence was found to challenge the novelty.
Finally, the versatility of LLMs has been instrumental in the development of the system, particularly for Zero-shot inference, reasoning-based activity extraction, and pseudo-labeling (see \ref{sec:methods}). This adaptability was crucial due to the lack of pre-existing models designed for such tasks. LLMs clearly open new opportunities for assisting large literature reviews in the pharmaceutical domain and, more broadly, across the biomedical domain.

\section{Conclusion}

Avoiding rediscoveries and dead-end paths is crucial in industrial antibiotic developments, saving time and resources. Yet, this process is itself resource-intensive, highlighting the need for semi-automatic reviewing. We present a practical application of LLMs to build an alert system that, given a list of organisms, flags evidence of previously reported activity from both the organism and chemical literature. We demonstrated the value of the system using 12 disclosed organisms and identified key factors: literature coverage, efficient natural products RE, synonym resolution and alert prioritization. The subset of the KG related to the negative hits, along with the code to reproduce the user interface and explore the results interactively is available at \url{Anonymous GitHub}.


\bibliography{custom}

\appendix

\section{Manual review and evaluation}
\label{sec:manual-review}
The review was conducted by a team of three experts (one biologist and two chemists) over several weeks (> 400 hours). In the process, they used PubMed, GBIF, CAS SciFinder \cite{gabrielson_scifinder_2018}, and LOTUS \cite{rutz_lotus_2022}. CAS SciFinder, a proprietary tool, facilitating the retrieval of scientific literature and patents related to chemical names and structures.

In the initial phase, reviewers examined literature associated with the target organisms, focusing on \texttt{OL}-evidence and chemicals produced by the organism (natural products). They also used GBIF to retrieve associated synonyms, and the LOTUS database to extend the search for natural products. As expected, few matches were found with the database, as the initial organism selection only involved weakly characterized organism. No filters were applied to the original studies, but, only secondary metabolites were retained and primary metabolites (those involved in growth, development or other essential pathways) were automatically excluded.

For each organism, reviewers compiled a list of compounds and primarily relied on SciFinder to explore associated literature and patents. Any evidence of antibiotic activity (growth inhibition, organism elimination, etc.) was considered as a hit, even if quantitative measurements (e.g., IC50 values) were not specified.

The reviewers emphasized that the first phase, identifying related natural products, is critical. Once compounds were identified, resources like SciFinder, alongside with expert knowledge, provide a detailed overview of the compounds' properties, literature, and associated patents. Nevertheless, the initial link between the organism and its chemical compounds remained often poorly documented. Finally, the goal is not to identify exhaustively all active molecules, rather, only identifying one or a few associated active compounds is sufficient to discard the organism.

\section{Activity Evidence Classification}
\label{app:AA-classifications}
\begin{figure*}[ht]
  \centering
  \includegraphics[width=0.8\textwidth]{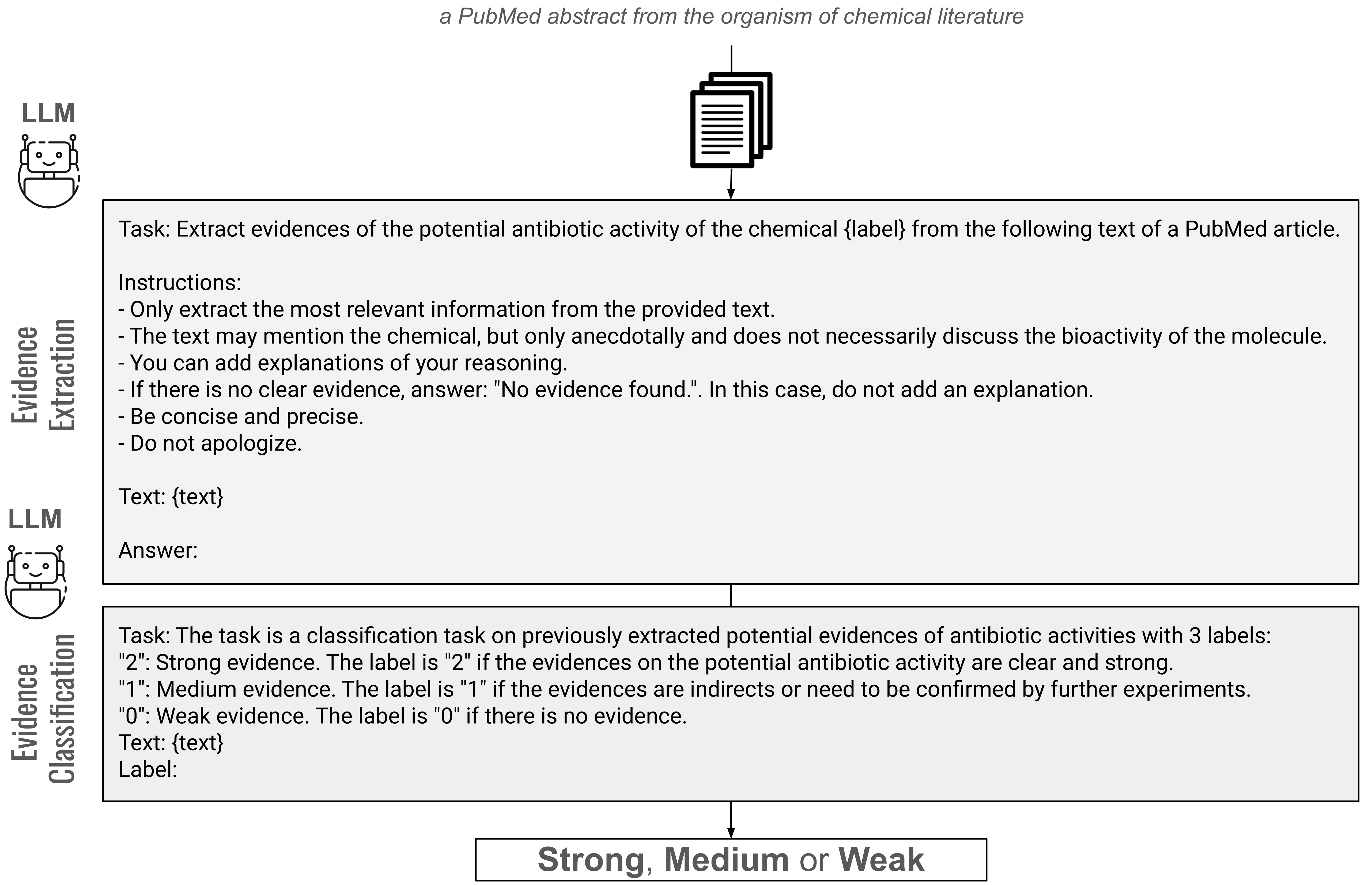}
  \caption{Schema of the prompting strategy for the extraction and classification of antibiotic activity evidence from the literature of chemicals (the strategy is equivalent for the literature of organisms).}
  \label{supp:AAprompts}
\end{figure*}

Concrete examples of \texttt{Strong}, \texttt{Medium} and \texttt{Weak} antibiotic evidence alerts, extracted using the prompting strategy described in Figure \ref{supp:AAprompts}.

\paragraph{Strong Activity Evidence: Cephalosporin C}

The following  evidence text has been extracted and classified as \texttt{Strong} from \texttt{PMID:4078571}: \textit{The text provides evidence that Cephalosporin C has antibacterial activity, particularly against Staphylococcus aureus. It is stated that Cephalosporin C has roughly the same activity as benzylpenicillin against several Gram-positive organisms and about one-eighth of the activity of benzylpenicillin against penicillin-sensitive strains of Staphylococcus aureus. Additionally, Cephalosporin C shows 4 to 8 times the activity of methicillin against penicillinase-producing staphylococcal strains. It also exhibits synergism in protection experiments in mice infected with a strong penicillinase-producing strain of Staphylococcus aureus when combined with benzylpenicillin.}". Here, the evidence of activity are supported by quantitative measurements.
 
\paragraph{Medium Activity Evidence: Cephalosporin C}
The following  evidence text has been extracted and classified as \texttt{Medium} from \texttt{PMID:22136576}: \textit{The evidence of the potential antibiotic activity of Acremostrictin is found in the statement "The new compound exhibited weak antibacterial activities." This suggests that Acremostrictin showed some level of antibacterial effect, although it was classified as weak.}. Here, the article only report weak antibacterial activity.

\paragraph{Weak Activity Evidence: Dipeptide delta-(L-alpha-aminoadipyl)-L-cysteine}
The following  evidence text has been extracted and classified as \texttt{Weak} from \texttt{PMID:6684424}: \textit{The text describes the biosynthesis of two compounds, the tripeptide delta-(L-alpha-aminoadipyl)-L-cysteinyl-D-valine and the dipeptide delta-(L-alpha-aminoadipyl)-L-cysteine, using a cell-free extract of Cephalosporium acremonium. However, it does not provide any information about the potential antibiotic activity of the dipeptide delta-(L-alpha-aminoadipyl)-L-cysteine. Therefore, there is No evidence found in this text to support the potential antibiotic activity of this chemical compound.}

\section{Filtering Classifiers}
\label{app:filtering-classifiers}
Considering the massive amount of literature to be processed for both \texttt{NPR} and activity extraction, it is essential to integrate a pre-filtering step to exclude out-of-scope references. It is also particularly essential for the RE step, which uses a decoder-only architecture where sending out-of-distribution abstracts (not mentioning any relations) lead to chaotic outputs. 

\subsection{\texttt{NPR} Filtering}
\label{sec:NPRfitlering}
\begin{figure*}[ht]
  \centering
  \includegraphics[width=0.8\textwidth]{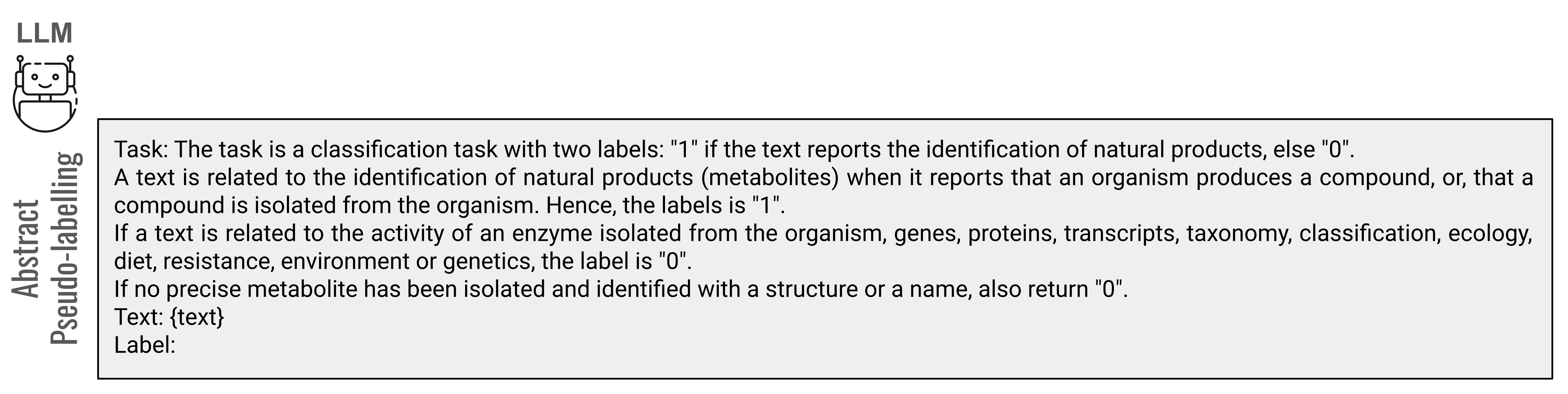}
  \caption{Prompt instructions for pseudo-labeling of natural products relationships.}
  \label{supp:prompt-pseudo-labels}
\end{figure*}

From the LOTUS database, we extracted the top-200 organism entities with the most associated relations and extracted 5k annotated abstracts, completed with 5k other abstracts not indexed in LOTUS. As LOTUS relations may not have been reported from the abstract (but from the full-text for instance) we annotated the dataset with LLM-generated pseudo-labels (prompt in Figure \ref{supp:prompt-pseudo-labels}). We trained a simple lexical Naive Bayes classifier and compared the performance against more complex transformer architecture BioBERT \cite{lee2020biobert} and SciBERT \cite{beltagy2019scibert} in Table \ref{tab:NPRfitlering}. 

\begin{table}[ht]
\centering
\begin{tabular}{lcccc}
\hline
\textbf{Model} & \textbf{Recall} & \textbf{Precision} & \textbf{F1}\\
\hline
Naive Bayes & 96.8 & 77.9 & 86.4 \\
BioBERT & 89.8 & 91.6 & 90.6 \\
SciBERT & 91.1 & 88.3 & 89.7 \\
\hline
\end{tabular}
\caption{Performance comparison of different models on \texttt{NPR} classification.}
\label{tab:NPRfitlering}
\end{table}

\subsection{Activity Filtering}
\label{ActivityFiltering}
While MeSH terms index articles in PubMed with relevant concepts such as \textit{Anti-Bacterial Agents}, most recent articles are not indexed, which can be critical for the alert system. Therefore, given the previously extracted top-200 organisms and their assocaited chemicals, we extracted their abstracts along with the MeSH annotations to build our dataset. We considered every article indexed with the concept \textit{Anti-Bacterial Agents} (or narrower in the hierarchy) as positive examples and the rest as negatives. From the total set, we re-sampled 5k positives and negatives. Similarly to \ref{sec:NPRfitlering} we trained a Naive Bayes classifier, BioBERT and SciBERT models (see Table \ref{tab:ActivityFiltering})

\begin{table}[ht]
\centering
\begin{tabular}{lcccc}
\hline
\textbf{Model} & \textbf{Recall} & \textbf{Precision} & \textbf{F1} & \textbf{F2} \\
\hline
Naive Bayes & 94.2 & 90.2 & 92.2 & 93.4 \\
BioBERT & 96.8 & 94.9 & 95.8 & 96.4 \\
SciBERT & 96.8 & 95.6 & 96.2 & 96.5 \\
\hline
\end{tabular}
\caption{Performance comparison of different models on \texttt{AA} classification.}
\label{tab:ActivityFiltering}
\end{table}

As expected, simple lexical approaches compete in practice with more complex transformers architecture, given the simplicity of the task. Indeed, in both cases, a keyword matching strategy is sufficient to efficiently classify the abstracts. We logically decided to use the simpler Naive Bayes Classifer in both cases

\section{Error analysis}
\label{sec:errorAnalysis}

This section provides a detailed error analysis on the 6 evidence the system failed to retrieve.

\textit{Acremonium butyri} - Orbuticin: While the chemical has been correctly extracted from the title of \texttt{PMID:8982351} the abstract and full-text of the article are not publicly available on PubMed, hence the system failed to extract the activity. The reported \texttt{Strong} evidence \textit{Acremonium butyri} in Figure \ref{fig:hits-alerts} actually refers to "Isoprenoids", which is a chemical family and not a single molecule. The \texttt{Strong} evidence is erroneously linked to articles reporting that the biosynthesis pathway for Isoprenoids is a target for many antibiotics.
    
\textit{Acrostalagmus luteoalbus} - T988 C: The RE model failed to extract the natural product from \texttt{PMID:35621985}. This relation is also not annotated in LOTUS.
    
\textit{Acrostalagmus luteoalbus} - Luteoalbusin A: The chemical has been correctly extracted from \texttt{PMID:35621985} but the activity information from \texttt{PMID:35621985} have not been extracted as only the abstract was processed.
    
\textit{Aspergillus calidoustus} - Strobilactone A: The article reporting the relation in LOTUS is not publicly available (\texttt{DOI:A10.7164/antibiotics.49.505})

\textit{Hypomyces aurantius} - Hypomycetin: The reference article identified by the reviewers (\texttt{DOI:10.3891/acta.chem.scand.51-0855}) is indexed in LOTUS. This article also describes the antifungal activity of Hypomycetin. However, since the article is not indexed in PubMed, the evidence of its activity has not been extracted.
    
\textit{Nectria inventa} - Verticillin B: The relation has correctly been identified in \texttt{PMID:31569621}, but the activity information from \texttt{PMID:31569621} have not been extracted as only the abstracts are processed.

\section{Screenshots of the User Interface}
\label{sec:screenshots}

\begin{figure*}[ht]
  \centering
  \includegraphics[width=1\textwidth]{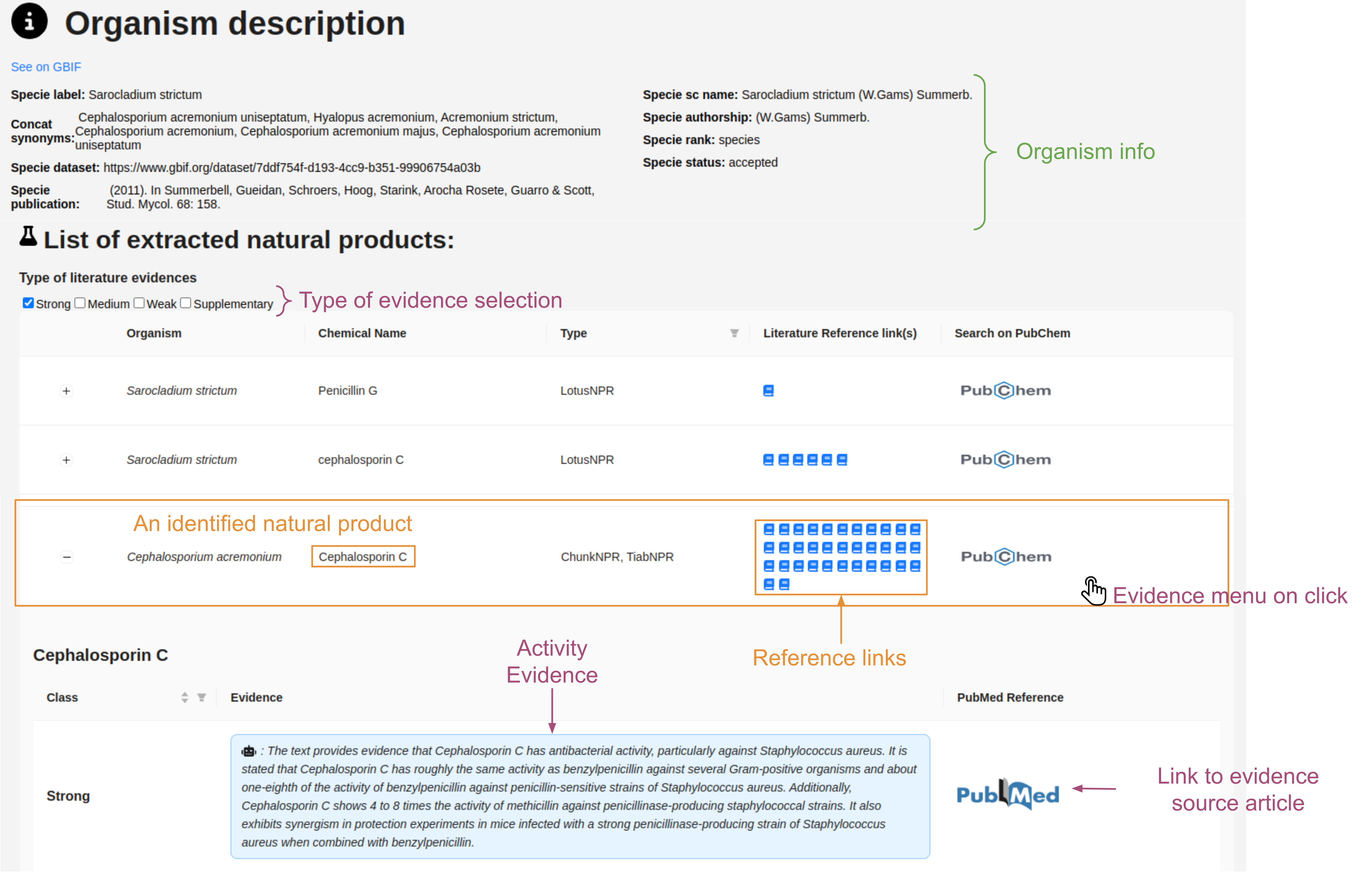}
  \caption{Screenshot of the \texttt{CL}-evidence alert panel for \textit{S. strictum}}
  \label{suppfig:screenshot-1}
\end{figure*}

\begin{figure*}[ht]
  \centering
  \includegraphics[width=1\textwidth]{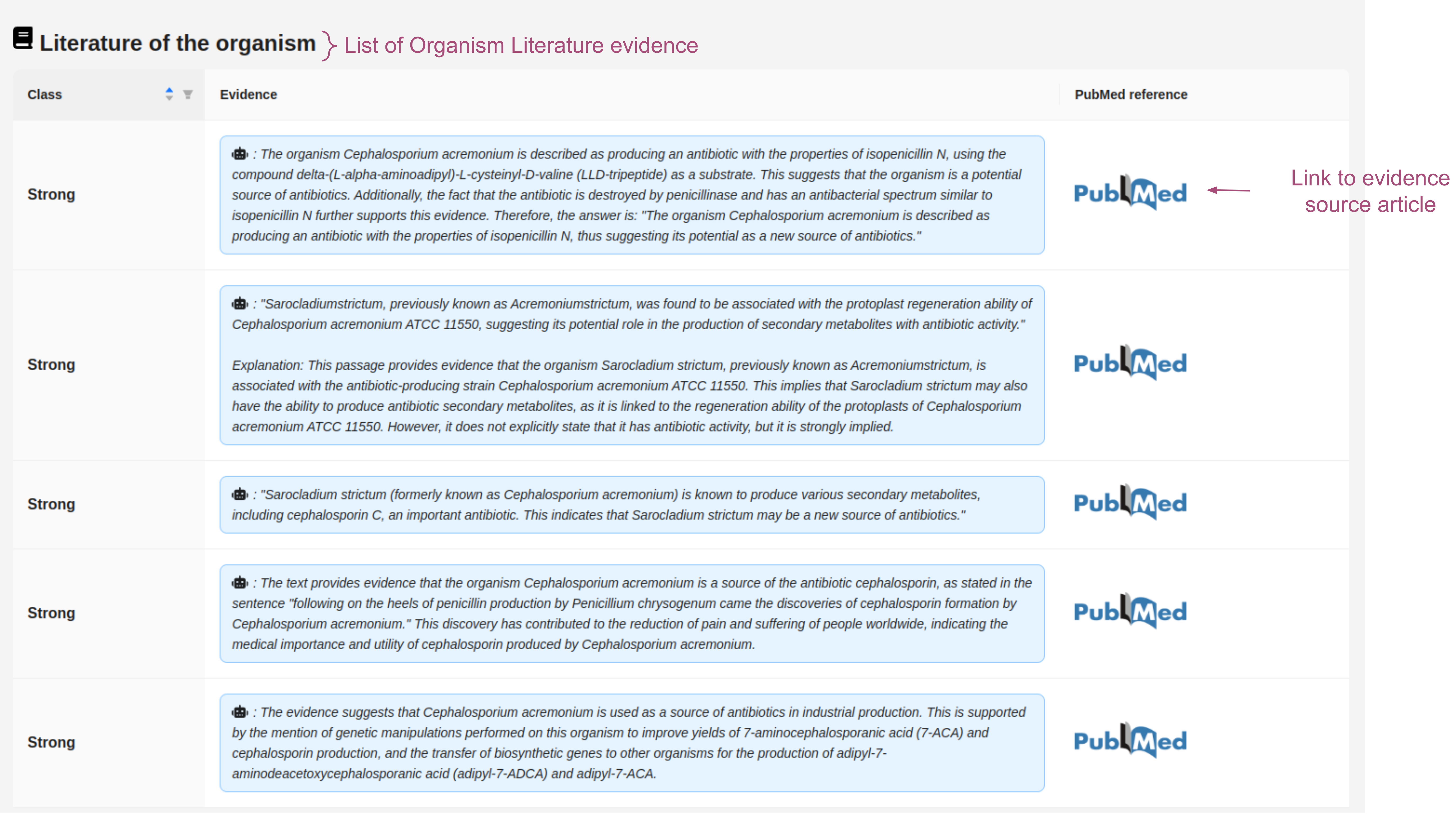}
  \caption{Screenshot of the \texttt{OL}-evidence alert panel for \textit{S. strictum}}
  \label{suppfig:screenshot-2}
\end{figure*}

\end{document}